\documentclass{article}
\usepackage{ijcai26}
\usepackage{times}
\usepackage{soul}
\usepackage{url}
\usepackage[hidelinks]{hyperref}
\usepackage[utf8]{inputenc}
\usepackage[small]{caption}
\usepackage{graphicx}
\usepackage{amsmath}
\usepackage{amssymb}
\usepackage{amsthm}
\usepackage{booktabs}
\usepackage{algorithm}
\usepackage{algorithmic}
\usepackage[switch]{lineno}

\title{Command-Space Counterfactual Explanations for Pareto-Conditioned Reinforcement Learning}

\author{
Joanikij Chulev$^1$\And
Hendrik Baier$^{1,2}$\\
\affiliations
$^1$ Centrum Wiskunde \& Informatica (CWI), Amsterdam\\
$^2$Eindhoven University of Technology\\
\emails
\{joanikij.chulev, hendrik.baier\}@cwi.nl   
}
\begin{document}

\maketitle

\begin{abstract}
Pareto Conditioned Networks learn multiple multi-objective reinforcement learning behaviours by conditioning a single policy on a desired return command. However, the local mapping from command and state to action remains opaque. We propose command-space counterfactual explanations for PCNs: given a fixed state, original command, and foil action, we search, in a black-box setting, for a minimally changed desired-return command under which the same trained policy would choose the foil. Our contributions are threefold. First, we formulate PCN explanations as return-command interventions, using a return-only PCN variant that avoids the added ambiguity of horizon-conditioning. Second, we adapt adversarial machine learning methods to reinforcement-learning explanations. Third, we introduce a boundary-seeded directional search that improves over purely local optimization in the command-action landscape, resulting in our proposed approach CF-ZOO. The resulting explanations are actionable and intuitively expressed in the user's own preferences: ``If your trade-off had shifted slightly towards X, the agent would have chosen Y.''

\end{abstract}

\section{Introduction}

Many sequential decision problems are inherently multi-objective: an agent may need to trade off aspects such as safety, efficiency or cost rather than optimize a single scalar reward. Multi-objective reinforcement learning (MORL) addresses this setting by using vector-valued rewards and learning policies that represent trade-offs among objectives \cite{roijers2013survey,hayes2022practical}. Such objective structure is also useful for XRL, where MORL has been identified as a route to contrastive explanation through explicit trade-offs among objectives \cite{milani2024xrl,dazeley2023broadxai}. The desired output is not one optimal policy but a set of Pareto-efficient behaviours, where improving one objective requires sacrificing another \cite{roijers2013survey}.

Pareto Conditioned Networks (PCNs) provide a compact way to represent such behaviours. Instead of learning a separate policy for each trade-off, a PCN learns one neural policy conditioned on a desired return and horizon \cite{reymond2022pareto}. At deployment, different desired-return commands can induce different behaviours from the same trained network. This makes PCNs attractive for decision support, but it does not make their decisions transparent: the command is interpretable, while the local mapping from state and command to action remains a neural black box.

Counterfactual explanations are well suited to this gap. In supervised learning, they explain how an input would need to change for a model to produce a different output \cite{wachter2018counterfactual,verma2024counterfactualreview}. In RL, however, counterfactuals are more complex, because decisions are sequential, stochastic, and connected to goals, plans, and future outcomes \cite{gajcin2024redefining}. Existing RL counterfactual work has therefore focused on visual state counterfactuals \cite{huber2023ganterfactualrl,samadi2024saferl}, reachable and certain state counterfactuals \cite{gajcin2024raccer}, diverse counterfactual action sequences \cite{gajcin2024acter}, and counterfactual modifications to policies themselves \cite{deshmukh2023counterpol}.

These approaches do not directly target the distinctive explanatory interface of a PCN: the desired-return command. For a PCN, a natural local question is: given the same state and the same trained policy, what minimal change to the command would make a difference? We propose an explanation, contrastive at the action level and interpretable at the objective level: it answers what desired-return trade-off would have made the same policy choose a different action in the same state. The result is directly actionable for the user. To compute these explanations, we propose \textsc{CF-ZOO}.

Methodologically, \textsc{CF-ZOO} adapts tools from adversarial and black-box optimization. Carlini--Wagner attacks formulate targeted behavioural change as an optimization problem balancing input proximity against a target-class loss \cite{carlini2017towards}. ZOO replaces back-propagation with finite-difference zeroth-order queries in black-box settings \cite{chen2017zoo}. Cheng et al. instead search over directions and scalar distances to decision boundaries to locate small hard-label perturbations \cite{cheng2019queryefficient}. Related work shows that adversarial perturbations can affect RL policies in domains such as algorithmic trading and autonomous lane changing \cite{faghan2020adversarial,zhang2023zerothorder}. We repurpose this optimization logic for explanation rather than attack: the perturbed object is the PCN command, and the goal is to expose how local action choice depends on the requested multi-objective trade-off.

A simple example illustrates the kind of explanation we seek. In Deep Sea Treasure, a submarine trades off treasure value against time cost. At one state, the submarine is one step above a nearby treasure of value \(5\), while moving right continues toward a larger treasure of value \(8\). The first component of the reward is the desired treasure value, and the second component is the desired time-cost return. Under the remaining command \(R_t=(8.000,-2.000)\), the trained PCN moves right. A user may ask: what would need to change for the same agent, in the same state, to move down instead? Our method returns \(R_{\mathrm{cf}}=(6.328,-1.704)\), with \(\delta R=(-1.672,0.296)\). This means that if the desired treasure return were reduced and the command placed slightly more pressure on finishing sooner, the same policy would move down to collect the nearer treasure. The explanation therefore reveals what the PCN has learned locally: the original action is tied to the command's preference for the larger treasure, not only to the submarine's current position. The user thus learns how to steer the policy toward the value-5 treasure as well.

Code is available at: \url{https://github.com/JoanikijChulev/Counterfactual-Explanations-for-Pareto-Conditioned-RL}.

\section{Background}

\subsection{Multi-Objective Reinforcement Learning}

A standard reinforcement learning problem is commonly formalized as a Markov decision process (MDP), where an agent observes a state, selects an action, receives a reward, and transitions to a next state \cite{sutton2018reinforcement}. MORL generalizes this setting by replacing the scalar reward with a vector reward. A multi-objective MDP can be written as
\[
    \mathcal{M}
    =
    \langle \mathcal{S},\mathcal{A},P,\mathbf{r},\gamma\rangle,
\]
where \(\mathcal{S}\) is the state space, \(\mathcal{A}\) is the action space, \(P(s'\mid s,a)\) is the transition function, \(\mathbf{r}(s,a,s')\in\mathbb{R}^m\) is an \(m\)-dimensional reward vector, and \(\gamma\in[0,1]\) is a discount factor \cite{roijers2013survey,hayes2022practical}. For a policy \(\pi\), the return is therefore vector-valued:
\[
    \mathbf{G}_t
    =
    \sum_{k=t}^{T}\gamma^{k-t}\mathbf{r}_k .
\]

MORL usually reasons in terms of Pareto dominance. A return vector \(\mathbf{x}\) dominates \(\mathbf{y}\) if it is at least as good in every objective and strictly better in at least one.

One way to solve a MORL problem is scalarization, where a utility function \(u:\mathbb{R}^m\rightarrow\mathbb{R}\) maps vector returns to a scalar objective. However, scalarization requires preferences to be specified and can miss parts of the Pareto front, especially under linear utilities \cite{roijers2013survey,hayes2022practical}. Multi-policy and conditioned-policy methods instead aim to represent many trade-offs within one learned system.  PCNs are one such conditioned-policy approach. Related approaches include preference-conditioned methods such as PD-MORL, which trains a single universal policy over preference space \cite{basaklar2023pdmorl}.

\subsection{Pareto Conditioned Networks and Our Return-Only Variant}
\label{subsec:pcn_return_only_variant}

PCNs learn a single policy network conditioned on a desired outcome \cite{reymond2022pareto}. In the original formulation, the policy is conditioned on the current state, a desired return vector, and a desired horizon:
\begin{equation}
    \pi_{\theta}(a_t \mid s_t, R_t^{\mathrm{des}}, h_t^{\mathrm{des}}),
\end{equation}
where \(R_t^{\mathrm{des}}\) specifies the desired multi-objective return and \(h_t^{\mathrm{des}}\) specifies the desired number of steps. During training, PCN stores experienced transitions together with their achieved returns and horizons, and learns to reproduce actions under the corresponding return--horizon commands \cite{reymond2022pareto}. Although horizon is part of the original PCN command, the original paper does not specifically show its role experimentally.

\begin{figure}[ht]
    \centering
    \includegraphics[width=0.95\linewidth]{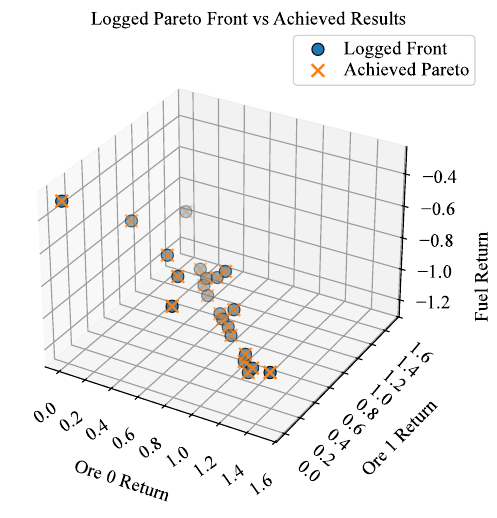}
    \caption{Pareto front recovered by the return-only PCN variant on Minecart. Blue points (O) show the final logged learning front; orange points (X) show rollout-achieved returns.}
    \label{fig:minecart_pareto_front}
\end{figure}

In this paper, we use a revised return-only variant:
\begin{equation}
    \pi_{\theta}(a_t \mid s_t, R_t^{\mathrm{des}}).
\end{equation}
During testing, we observed that horizon conditioning was not always a reliable control signal. Even when the desired return command corresponded to a Pareto-optimal point, changing only the desired horizon could sometimes produce unstable or unintuitive action choices.
This suggests that the horizon input may not have consistently encoded useful temporal guidance for the policy.
Using the return-only formulation, we trained PCN agents and recovered the expected Pareto-optimal trade-off structure in our environments.
For example, the learned Minecart environment front contained more Pareto-optimal solutions than reported for the original implementation \cite{reymond2022pareto,abels2019dynamic}. Figure~\ref{fig:minecart_pareto_front} shows the Pareto front recovered by our return-only variant on Minecart.

Furthermore, our explanation objective is to understand how desired-return commands affect action choice. Keeping the horizon would require counterfactuals over both return and time,
\[
    (R,h) \mapsto (R+\delta_R, h+\delta_h),
\]
whereas the return-only formulation yields the simpler explanation problem
\[
    R \mapsto R+\delta_R.
\]

For a fixed state \(s\), the relevant object is therefore the local action distribution induced by the return command. Let \(\pi_\theta(s,R) \in \mathbb{R}^{|\mathcal{A}|}\) denote the action probabilities produced by the policy. The greedy action is
\begin{equation}
a^\star(s,R) = \arg\max_{a\in \mathcal{A}} \pi_\theta(a\mid s,R).
\end{equation}
A command-space counterfactual then asks for a small perturbation \(\delta_R\) such that the preferred action changes. This makes the desired return an interpretable intervention point.

\subsection{Counterfactual Explanations in Reinforcement Learning}

A counterfactual explanation is usually defined relative to an original input, an original model output, and a desired alternative output \cite{wachter2018counterfactual,verma2024counterfactualreview}. In supervised learning, counterfactual search is often formulated as an optimization problem combining validity, proximity, sparsity, plausibility, and sometimes diversity \cite{mothilal2020dice,dandl2020moc}.

In RL, these desiderata are not sufficient. RL decisions occur inside temporally extended interaction loops, so a counterfactual state that is close in feature space can still be unreachable from the original state under the environment dynamics \cite{gajcin2024redefining}. Stochasticity adds another complication: a counterfactual may produce the desired action or outcome in one rollout but fail under nearby random transitions. 

Existing work addresses these issues in several ways. Olson et al. generate counterfactual state explanations for visual RL agents using generative models \cite{olson2021counterfactual}. Tsirtsis et al. formulate counterfactual explanations for sequential decision making under uncertainty as alternative action sequences or policies \cite{tsirtsis2021counterfactual}. RACCER introduces RL-specific properties such as reachability, stochastic certainty, and fidelity \cite{gajcin2024raccer}, while ACTER extends the focus to diverse counterfactual action sequences \cite{gajcin2024acter}. COViz explains local RL decisions by comparing the outcome of the chosen action with a counterfactual action outcome \cite{amitai2024coviz}. 

GANterfactual-RL and SAFE-RL focus on visual counterfactuals for deep RL policies, where the goal is to alter high-dimensional observations in ways that change the agent's action while preserving plausibility \cite{huber2023ganterfactualrl,samadi2024saferl}.

PCN command-space counterfactuals differ from these approaches. They do not modify pixels, symbolic state variables, or past action sequences. They ask how the user's requested trade-off would need to change for a different local action to become preferred. Such explanations are especially natural for MORL because the focus is on objectives.

\section{Command-Space Counterfactual Search}
\label{sec:command_space_cf}

Let \(s_t\) be the state to explain, \(R_t\in\mathbb{R}^{m}\) the remaining desired-return command, \(a^\star\) the greedy action selected by the PCN, and \(a_f\neq a^\star\) a valid foil action. The policy is queried only as a black box: for a command \(R\), it returns log-probabilities \(\ell_a(s_t,R)\). The valid action set at \(s_t\) is denoted by \(\mathcal{A}(s_t)\).

A command-space counterfactual is a command
\begin{equation}
    R_{cf}=R_t+\delta
\end{equation}
such that the foil becomes the greedy action:
\begin{equation}
    a_f
    =
    \arg\max_{a\in\mathcal{A}(s_t)}
    \ell_a(s_t,R_{cf}).
    \label{eq:cf_greedy_condition}
\end{equation}
The resulting explanation has the form: the policy selected \(a^\star\) under \(R_t\), but would have selected \(a_f\) under \(R_{cf}\). The state and policy are held fixed; only the desired-return command is changed.

\subsection{Optimization Objective}

Our loss adapts the targeted margin structure used in Carlini--Wagner-style attacks and ZOO \cite{carlini2017towards,chen2017zoo}. The difference is semantic: adversarial attacks perturb an input to induce misclassification, whereas we perturb the PCN command to expose how the local action preference depends on the requested multi-objective return.

For a foil \(a_f\), define the target margin:
\begin{equation}
    M_f(R)
    =
    \ell_{a_f}(s_t,R)
    -
    \max_{a\in\mathcal{A}(s_t),\,a\neq a_f}
    \ell_a(s_t,R).
    \label{eq:target_margin}
\end{equation}
The foil is greedy when \(M_f(R)\geq 0\). We use a margin parameter \(\kappa\geq0\) and require \(M_f(R)\geq\kappa\), so that the foil wins by a nonzero margin when strict flips are desired. Equivalently, define the hinge term
\begin{equation}
    H_f(R)
    =
    \max\left\{
        \kappa - M_f(R),
        0
    \right\}.
    \label{eq:hinge}
\end{equation}
This term is zero exactly when the foil action beats all other valid actions by at least \(\kappa\).

The black-box objective optimized by the explainer is
\begin{equation}
    \mathcal{L}(\delta)
    =
    \|\delta\|_2^2
    +
    c\,H_f\!\left(\Pi_{\mathcal{C}}(R_t+\delta)\right),
    \label{eq:pcn_zoo_objective}
\end{equation}
where \(c>0\) controls the strength of the targeted flip penalty and \(\Pi_{\mathcal{C}}\) clips commands to the feasible box. The first term selects small command changes. The second term enforces the foil-action preference.

\subsection{Boundary-Seeded Directional Search}
\label{subsec:boundary_seeded_search}

A purely local C\&W/ZOO-style search can fail when the command-action landscape is flat, non-convex, or when the current command is far from the foil region. In that case, finite-difference updates around \(R_t\) may only observe weak local changes in the target margin and can converge to an uninformative local solution that never makes the foil action greedy. We therefore add directional boundary seeding. The purpose is not to replace the C\&W/ZOO objective, but to provide it with a better initial perturbation direction.

This idea is adapted from query-efficient hard-label black-box attacks. Cheng et al. reformulate hard-label attack search by optimizing over directions rather than directly over perturbed inputs \cite{cheng2019queryefficient}. For an input \(x_0\), classifier \(f\), and target class \(t\), their targeted boundary-distance objective is
\begin{equation}
    g(\theta)
    =
    \min_{\lambda>0}\ \lambda
    \quad\text{s.t.}\quad
    f\!\left(x_0+\lambda\frac{\theta}{\|\theta\|_2}\right)
    =
    t.
    \label{eq:cheng_targeted_original}
\end{equation}
Inspired by these attacks, we use a directional command-space search: candidate directions are evaluated by testing their feasible endpoint, and successful directed perturbations are binary-refined back toward the original command to approximate the nearest successful perturbation along that direction.

The seeded directional search assumes access to the training-time Pareto archive. Thus, we check for counterfactuals in the directions of all found Pareto front solutions, relying on them as behavioral heuristics. Let \(\mathcal{F}=\{R^{\mathrm{front}}_j\}_{j=1}^{N}\) denote the logged Pareto-front returns. At timestep \(t\), the current command \(R_t\) is the remaining part of the originally selected command \(R_{\mathrm{des}}\). Hence the return already collected is
\[
    R_t^{\mathrm{col}} = R_{\mathrm{des}} - R_t .
\]
Each front point is converted into the remaining command that would still be
needed to reach it:
\[
    \tilde{R}_j = R_j^{\mathrm{front}} - R_t^{\mathrm{col}} .
\]
We query the PCN at each \(\tilde{R}_j\), compute the foil margin
\(M_f(\tilde{R}_j)\), and select
\[
    j^\star = \arg\max_j M_f(\tilde{R}_j).
\]

This candidate is used only to infer a coarse direction of useful
command change. Specifically, the directional sign is computed from

\[
d = \tilde R_{j^\star} - R_t .
\]

For component \(i\), if \(d_i>0\), subsequent ray search is restricted
to increasing that command component; if
\(d_i<0\), it is restricted to decreasing that component; and if
\(d_i=0\), no directional restriction is imposed on that component.

After inferring the directional prior, we sample unit candidate directions as
\begin{equation}
    v \sim \mathcal{N}(0,I_d),
    \qquad
    u = \frac{v}{\lVert v\rVert_2},
    \label{eq:gaussian_direction}
\end{equation}
where \(I_d\) is the \(d \times d\) identity matrix, so \(v\) is a
\(d\)-dimensional standard Gaussian vector. For each \(u\), we search along
the command-space ray, where \(D=\operatorname{diag}(R_{\max}-R_{\min})\) scales the command dimensions
\begin{equation}
    R(\alpha;u) = R_t + \alpha D u,
    \qquad
    \alpha \geq 0.
    \label{eq:command_ray}
\end{equation}
The endpoint of the perturbation is the largest feasible \(\alpha\) that remains inside both the reward bounds and the directional bounds inferred from the Pareto-front heuristic. If this endpoint does not satisfy
\begin{equation}
    M_f(R(\alpha;u))\geq\kappa,
\end{equation}
the point is discarded. If the endpoint succeeds, we binary-search back toward \(R_t\) to remove unnecessary perturbation magnitude. All successful candidates are retained and compared using the scaled command distance
\begin{equation}
    d_D(R,R_t)
    =
    \left\|D^{-1}(R-R_t)\right\|_2.
    \label{eq:scaled_boundary_distance}
\end{equation}
The closest successful ray candidate becomes the seed for the subsequent ZOO coordinate refinement. If no ray succeeds, the method falls back to local ZOO from the original command. Thus, the boundary-seeding phase supplies a global directional guess, while the ZOO phase performs local black-box refinement around the best found candidate.

\subsection{ZOO Perturbation Refinement}

After directional seeding, we refine the best seed using ZOO-style zeroth-order coordinate optimization \cite{chen2017zoo}. ZOO replaces back-propagation with finite-difference queries to the target model. In our setting, the target model is the PCN queried at the fixed state \(s_t\), and the optimized variable is the command perturbation \(\delta\).

At each iteration, a coordinate \(i\in\{1,\ldots,d\}\) is sampled. The coordinate derivative of Eq.~\eqref{eq:pcn_zoo_objective} is estimated by central finite differences:
\begin{equation}
    \widehat{\nabla_i \mathcal{L}}(\delta)
    =
    \frac{
        \mathcal{L}(\delta+he_i)
        -
        \mathcal{L}(\delta-he_i)
    }{2h},
    \label{eq:finite_difference}
\end{equation}
where \(e_i\) is the \(i\)-th basis vector and \(h\) is the finite-difference step. Each coordinate update therefore uses two score queries for the finite-difference estimate and one additional query to evaluate the updated command. We use ZOO-ADAM for optimization of coordinate updates.

\section{Worked Examples}
\label{sec:worked_examples}

\begin{figure}[ht]
    \centering
    \includegraphics[width=\columnwidth]{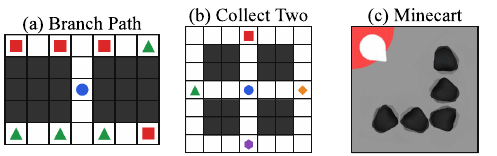}
    \caption{(a) Branch-Path Environment - Custom. (b) Collect-Two Environment - Custom. (c) Minecart Environment - Classic.}
    \label{fig:worked_envs}
\end{figure}

Figure~\ref{fig:worked_envs} shows the three environments used for qualitative inspection. The purpose is not only to show that an action flip occurs, but also to show how the result can be communicated as a human-facing explanation of the agent's behaviour.

\subsection{Branch-Path}
\label{subsec:branch_path_worked_example}

Branch-Path is a custom \(5\times 7\) two-objective grid with a central start state and two narrow branches. Collectibles of type \(A\) (red square) give reward \((1,0)\), while collectibles of type \(B\) (green triangle) give reward \((0,1)\). Invalid moves are masked, so the search only compares feasible actions. There are three Pareto-front points, due to an episode step limit of 13 steps.
We select
\[
    R_t=(3,1),
\]
which prioritizes objective \(A\). At timestep \(t=1\), the remaining command is still \(R_t=(3,1)\). The valid actions are \texttt{up} and \texttt{down}. The PCN assigns probability \(0.995\) to \texttt{up} and \(0.005\) to \texttt{down}, so
\[
    a^\star=\texttt{up}.
\]
We choose the foil
\[
    a_f=\texttt{down}.
\]

The search returns
\[
    R_{cf}=(0.2953,1.464),
    \qquad
    \delta_R=(-2.7047,0.464).
\]
Under \(R_{cf}\), the greedy action changes from \texttt{up} to \texttt{down}. The intuitive explanation here is:
\emph{At this state, the agent goes up because the current command strongly asks for objective \(A\). If the user instead asked for much less \(A\) and somewhat more \(B\), the same agent in the same state would go down. The upward action is therefore not arbitrary: it is tied to the command's preference for \(A\).}

\subsection{Collect-Two}
\label{subsec:collect_two_worked_example}

Collect-Two is a custom \(7\times 7\) four-objective grid. The agent starts at the center, with objective \(A\) (red) above, \(B\) (green) left, \(C\) (yellow) right, and \(D\) (purple) below. The task terminates after two objectives are collected. The first collected objective gives reward \(1.0\), and the second gives reward \(0.8\). Thus, the Pareto front contains 12 points. At timestep \(t=0\), the agent is at the center state with command
\[
    R_t=(0,0,0.8,1).
\]
This command values objective \(C\), but values objective \(D\) most. The PCN assigns probabilities \(0.145\), \(0.128\), \(0.132\), and \(0.595\) to \texttt{right}, \texttt{up}, \texttt{left}, and \texttt{down}, respectively. Therefore,
\[
    a^\star=\texttt{down}.
\]
We choose
\[
    a_f=\texttt{right}.
\]

The search returns
\[
    R_{cf}=(0,0,0.8,0.8437),
    \qquad
    \delta_R=(0,0,0,-0.1563).
\]
Under \(R_{cf}\), we are making \texttt{right} greedy. We can explain this as:
\emph{At the center, the agent goes down because objective \(D\) is requested slightly more strongly than objective \(C\). If the desired return for \(D\) were reduced and made similar to \(C\), the agent would instead go right toward \(C\). If the agent values them both similarly, it would go right. This is a sparse explanation: only one objective in the command needs to change.}

\subsection{Minecart}
\label{subsec:minecart_worked_example}

Minecart is a continuous-state MORL benchmark in which a cart moves through a two-dimensional map, collects ore, and trades off ore objectives against fuel consumption \cite{abels2019dynamic}. This example tests whether command counterfactuals remain interpretable outside small discrete grids.

At timestep \(t=5\), the state is
\[
    s_t=[0.4876,0.1991,0.03,0.0872,0.9962,0,0],
\]
corresponding to a cart near \((x,y)=(0.488,0.199)\), moving slowly, approximately facing east, with empty cargo. The command is
\[
    R_t=(0.28,1.22,-0.96),
\]
where the first two dimensions are ore objectives and the third is fuel-related. The PCN assigns probabilities \(0.999860\), \(0.000005\), and \(0.000134\) to \texttt{Left}, \texttt{Right}, and \texttt{None}, respectively, so
\[
    a^\star=\texttt{Left}.
\]
We choose
\[
    a_f=\texttt{Right}.
\]

The search returns
\[
    R_{cf}=(0.3476,1.22,-0.96),
    \qquad
    \delta_R=(0.0676,0,0).
\]
Under \(R_{cf}\), \texttt{Right} becomes greedy, which is consistent with a sensible learned trade-off. This can be explained as:
\emph{Given the state, the agent strongly turns left under the original command. If the user requested slightly more of the first ore objective, while leaving the second ore objective and fuel command unchanged, the same agent would instead turn right. Increasing the desired amount of ore 0 makes the policy stop steering toward the nearby mine containing only ore 1 and instead steer clockwise toward mines that contain ore 0.}

\section{Experimental Results}
\label{sec:experiments}
We use a mixture of custom diagnostic environments and standard MORL benchmarks, several of which are provided through MO-Gymnasium and MORL-Baselines \cite{felten2023mogymnasium}. Deep Sea Treasure (DST) is a classic MORL grid benchmark in which a submarine trades off treasure value against a time penalty \cite{vamplew2011empirical}. Breakable-Bottles is a low-impact MORL benchmark where the agent must deliver bottles while accounting for time and potential environmental side effects \cite{vamplew2021potential}. Resource-Gathering is a grid-world task in which the agent collects resources such as gold and gems while facing enemy risk, originally introduced by Barrett and Narayanan \cite{barrett2008learning}. Fruit-Tree Navigation is a tree-structured MORL benchmark in which each path leads to a fruit with a multi-objective nutrient vector, testing generalization over larger reward spaces \cite{yang2019generalized}. Reward-Line is our custom diagnostic environment: the agent moves in a grid to a terminal row where the terminal column defines a linear two-objective trade-off between \((1,0)\) and \((0,1)\).

\subsection{White-Box Baseline}
\label{subsec:white_cw_baseline}

We first test whether the command counterfactual problem can be solved by a strong local optimizer when model internals are available. The baseline, denoted \textsc{White-CW}, adapts the Carlini--Wagner targeted attack objective to PCN command vectors and optimizes it through the Adversarial Robustness Toolbox (ART) \cite{carlini2017towards,nicolae2018art}. Unlike our method, this baseline has white-box access to the model and can use internal gradients. A comparison to this baseline allows us to test how many of the failures are due to black-box access.

Our method, denoted \textsc{CF-ZOO}, uses the same foil-validity condition but first performs boundary-seeded directional search before zeroth-order local refinement. A case is successful if the method finds a feasible command \(R_{cf}\) for which the selected foil action becomes greedy.

\begin{table}[ht]
\centering
\footnotesize
\setlength{\tabcolsep}{1.8pt}
\begin{tabular*}{\columnwidth}{@{\extracolsep{\fill}}l l r r r r@{}}
\toprule
Env. & Method & Cases & Succ. & Rate [95\% CI] & Dist. \\
\midrule
Branch & \textsc{White-CW} & 35 & 7  & 20.0 [10.0,35.9] & 0.270 \\
Branch & \textsc{CF-ZOO}  & 35 & 25 & 71.4 [54.9,83.7] & 0.416 \\
B-Bottles & \textsc{White-CW} & 8 & 2  & 25.0 [7.1,59.1] & 0.017 \\
B-Bottles & \textsc{CF-ZOO}  & 8 & 6  & 75.0 [40.9,92.9] & 0.030 \\
Collect & \textsc{White-CW} & 136 & 29  & 21.3 [15.3,28.9] & 0.514 \\
Collect & \textsc{CF-ZOO}  & 136 & 136 & 100.0 [97.3,100.0] & 0.444 \\
DST & \textsc{White-CW} & 136 & 14 & 10.3 [6.2,16.5] & 0.014 \\
DST & \textsc{CF-ZOO}  & 136 & 50 & 36.8 [29.1,45.1] & 0.152 \\
Fruit & \textsc{White-CW} & 84 & 83 & 98.8 [93.6,99.8] & 0.130 \\
Fruit & \textsc{CF-ZOO}  & 84 & 84 & 100.0 [95.6,100.0] & 0.121 \\
Minecart & \textsc{White-CW} & 240 & 116 & 48.3 [42.1,54.6] & 0.098 \\
Minecart & \textsc{CF-ZOO}  & 240 & 208 & 86.7 [81.8,90.4] & 0.153 \\
R-Gather & \textsc{White-CW} & 10 & 0 & 0.0 [0.0,27.8] & -- \\
R-Gather & \textsc{CF-ZOO}  & 10 & 4 & 40.0 [16.8,68.7] & 1.186 \\
R-Line & \textsc{White-CW} & 148 & 71 & 48.0 [40.1,56.0] & 0.380 \\
R-Line & \textsc{CF-ZOO}  & 148 & 124 & 83.8 [77.0,88.9] & 0.363 \\
\midrule
All & \textsc{White-CW} & 797 & 322 & 40.4 [37.0,43.8] & -- \\
All & \textsc{CF-ZOO}  & 797 & 637 & 79.9 [77.0,82.6] & -- \\
\bottomrule
\end{tabular*}
\caption{Comparison of our perturbation method with a white-box C\&W baseline. Success-rate intervals are Wilson score \(95\%\) confidence intervals. Dist. denotes the mean scaled \(\ell_2\) distance over successful counterfactuals.}
\label{tab:experimental_results}
\end{table}

\textsc{CF-ZOO} succeeds in \(637/797\) cases (\(79.9\%\)), compared with
\(322/797\) (\(40.4\%\)) for \textsc{White-CW}. Except in Fruit-Tree,
boundary seeding substantially improves validity, indicating that white-box
access alone does not overcome poor local initialization. Distances are
averaged only over successful cases, so larger \textsc{CF-ZOO} distances can
reflect additional, harder cases that \textsc{White-CW} does not solve.
 For example, in DST it succeeds in only \(14/136\) cases with mean distance \(0.014\), whereas \textsc{CF-ZOO} succeeds in \(50/136\) cases with mean distance \(0.152\). The larger distance for \textsc{CF-ZOO} reflects that it reaches foil regions that the local white-box optimizer misses.
These results support the boundary-seeding design: even with model internals available, local C\&W-style optimization often fails to cross the action boundary, while \textsc{CF-ZOO} more reliably reaches feasible foil regions.

\subsection{Command-Space Landscape Exploration}
\label{subsec:landscape_visualizations}

\begin{figure*}[t]
    \centering
    \includegraphics[width=\textwidth]{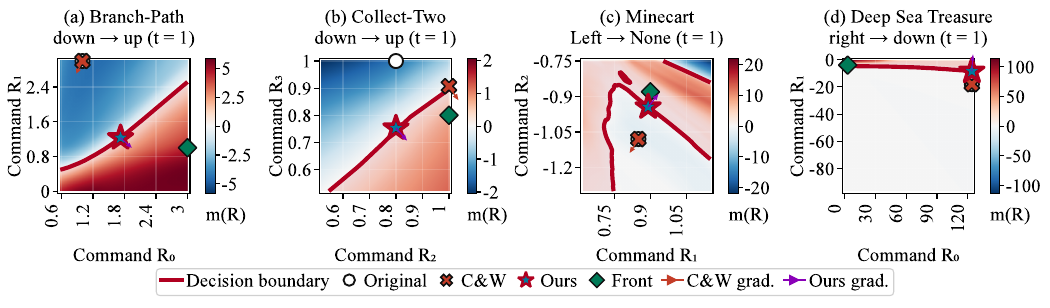}
    \caption{Command-space decision landscapes for four qualitative showcase
    scenarios. The title explains the flip, which return was chosen, and the
    timestep of the rollout. We mark the original command (circle) and the C\&W (cross), our CF-ZOO (star), and nearest Pareto-front (diamond) counterfactuals, with arrows for the local margin gradient. Axes show true values in rewards.}
    \label{fig:command_landscapes}
\end{figure*}

A closer look at the command space helps explain why the two methods behave differently. Figure~\ref{fig:command_landscapes} shows four representative
two-dimensional slices of the command space. In panels (a), (c), and (d),
\textsc{White-CW} finds no valid counterfactual, so its marker overlaps the original command \(R_t\); only in panel (b) does it reach the validity margin. The arrows show the local direction of increasing foil margin and are diagnostic. In panel (a), this direction is poorly aligned with a nearby valid region, while panel (c) shows that even an apparently useful local direction does not guarantee success. These landscape figures therefore illustrate how such landscapes can limit purely local optimization. By screening nonlocal rays, \textsc{CF-ZOO} reaches a foil-valid region in all four cases.

\subsection{Combinatorial Stress Test}
\label{subsec:stress_test}
We further evaluate our implementation in a stress test. Instead of selecting a small set of representative counterfactual queries, we generate explanation cases across available Pareto-front commands, sampled timesteps, and all valid foil actions at each selected state. This produces a full evaluation of whether the method performs under many decision contexts. The white-box comparison (Table~\ref{tab:experimental_results}) used the same procedure restricted to fewer front commands, early timesteps and fewer foils so its cases are a subset of Table~\ref{tab:stress_test}.

\begin{table}[ht]
\centering
\footnotesize
\setlength{\tabcolsep}{2.0pt}
\begin{tabular*}{\columnwidth}{@{\extracolsep{\fill}}l r r r r@{}}
\toprule
Environment & Cases & Succ. & Rate [95\% CI] & Dist. \\
\midrule
Branch-Path & 44 & 30 & 68.2 [53.4,80.0] & 0.466 \\
B-Bottles & 8 & 6 & 75.0 [40.9,92.9] & 0.030 \\
Collect-Two & 152 & 152 & 100.0 [97.5,100.0] & 0.493 \\
DST & 243 & 70 & 28.8 [23.5,34.8] & 0.641 \\
Fruit-Tree & 672 & 672 & 100.0 [99.4,100.0] & 0.231 \\
Minecart & 1846 & 1341 & 72.6 [70.6,74.6] & 0.360 \\
R-Gather & 29 & 10 & 34.5 [19.9,52.7] & 1.131 \\
Reward-Line & 201 & 141 & 70.1 [63.5,76.0] & 0.525 \\
\midrule
Total & 3195 & 2422 & 75.8 [74.3,77.3] & -- \\
\bottomrule
\end{tabular*}
\caption{Stress test of \textsc{CF-ZOO}. Each case corresponds to a state, command, and valid foil action. Success-rate intervals are Wilson score \(95\%\) confidence intervals. Dist. denotes the mean scaled \(\ell_2\) distance over successful counterfactuals.}
\label{tab:stress_test}
\end{table}

Across \(3{,}195\) cases, \textsc{CF-ZOO} succeeds in \(2{,}422\)
(\(75.8\%\)). The method succeeds on all Collect-Two and Fruit-Tree cases, and achieves high success on Minecart and Reward-Line, indicating that the search scales beyond small diagnostic grids and remains effective in continuous-state and structured multi-objective settings. The lower success rates in DST and Resource-Gathering suggest that some environments contain local decisions where the foil action is difficult to induce through command changes alone. In these cases, the policy may be strongly constrained by the state or the desired foil may lie outside the learned behavioural support.
The mean scaled distances show that the magnitude of successful explanations varies substantially across environments. Overall, the stress test shows that boundary-seeded command search can recover counterfactual explanations for most generated foil queries while also exposing where the PCN does not provide reliable command-level control. 

To determine whether these failures reflect a limitation of the search or
the absence of a counterfactual among the evaluated commands, we
exhaustively evaluated a dense finite grid over the command space at each
failing case. A foil for which no grid command flips the action is treated
as heuristic evidence that the policy may not have learned that behaviour.
Table~\ref{tab:cfzoo_taxonomy} decomposes \textsc{CF-ZOO}'s outcomes. Of
all cases, \(78.7\%\) are solved, while \(16.6\%\) are grid-infeasible
foils---actions that no evaluated grid command flips. Restricting to the known-feasible cases,
for which \textsc{CF-ZOO} or the grid found at least one valid
counterfactual, \textsc{CF-ZOO} recovers \textbf{\(94.4\%\)}. The remaining
\(5.6\%\) are search failures that persist even when the search budget is
high, indicating a small set of known-feasible counterfactuals that the
method struggles to locate.

\begin{table}[ht]
\centering
\footnotesize
\setlength{\tabcolsep}{1.8pt}
\begin{tabular*}{\columnwidth}{@{\extracolsep{\fill}}l r r r@{}}
\toprule
Outcome & Cases & \% all [95\% CI] & \% feasible [95\% CI] \\
\midrule
Success$^{\dagger}$      & 2515 & 78.7 [77.3,80.1] & \textbf{94.4} [93.5,95.2] \\
Grid-infeasible foil     &  531 & 16.6 [15.4,18.0] & -- \\
Search-limited failure   &  149 &  4.7 [4.0,5.5]   & 5.6 [4.8,6.5] \\
\bottomrule
\end{tabular*}
\caption{CF-ZOO diagnosis. Feasibility is assessed using a grid with 200 values per command dimension, yielding \(200^d\) commands for \(R\in\mathbb{R}^d\). Percentages in the \% feasible column use only the feasible cases. \(^{\dagger}\)Includes 93 additional cases recovered by increasing the budget to 100k. Brackets denote 95\% Wilson confidence intervals.}
\label{tab:cfzoo_taxonomy}
\end{table}

\section{Discussion and Scope}

The output should be interpreted locally. A successful counterfactual means that, at state \(s_t\), the same trained PCN would prefer \(a_f\) under command \(R_{cf}\). It does not imply that \(R_{cf}\) is Pareto-optimal. Conversely, failure may indicate optimizer failure, an unsupported foil, or a policy that is locally insensitive to the desired-return command.

Gajcin and Dusparic argue that RL counterfactuals cannot be imported directly from supervised learning because RL decisions are sequential and temporally embedded \cite{gajcin2024redefining}. RACCER addresses this by searching for an action sequence that reaches a counterfactual state where the desired action is likely under the policy \cite{gajcin2024raccer}. This answers the question of how an agent could reach another state in which it would choose the foil action. Our question is different. We ask why the PCN did not choose \(a_f\) at the original state \(s_t\). Moving the agent to another state can be operationally reachable but explanatorily indirect. We therefore intervene on the PCN command rather than the environment state. Since the desired-return command is the user-facing trade-off input, \(R_{cf}\) is actionable by construction: it can be issued directly to the same policy at the same state. This does not guarantee that the resulting return is achievable, but it makes the counterfactual intervention itself directly implementable.

The seven desiderata therefore specialize differently in our setting \cite{gajcin2024redefining}. \emph{Validity} is the margin condition \(M_f(R_{cf})\geq\kappa\), meaning the foil becomes preferred by the PCN. \emph{Proximity} is the scaled command distance \(\|D^{-1}(R_{cf}-R_t)\|_2\), which keeps the requested trade-off close to the original command. \emph{Actionability} is command actionability: the user can directly issue the counterfactual command. \emph{Sparsity} corresponds to changing few objective dimensions, although our current objective emphasizes small scaled distance rather than an explicit \(\ell_0\) penalty. In practice, the explanations are usually sparse, changing 1 or 2 entries. \emph{Data-manifold closeness} is approximated by restricting commands to feasible bounds and using logged Pareto-front commands as behavioural anchors. \emph{Causality} is local and interventional: the state and trained model are held fixed while only the command is changed. \emph{Recourse} is immediate: replacing the directly settable command $R_t$ with $R_{cf}$ realizes the counterfactual.

\section{Conclusion and Future Work}
\label{sec:conclusion}

This paper introduces desired-return counterfactual explanations for command-conditioned MORL policies, such as PCNs, and proposes \textsc{CF-ZOO} to compute them. For a fixed policy and state, each explanation identifies how the requested trade-off would need to change to induce a specified foil action. Although currently PCN-specific, the approach could extend to other MORL or goal-conditioned agents with user-controllable preference, goal, or utility-conditioning inputs, such as PD-MORL agents \cite{basaklar2023pdmorl}.

The method also has limitations. Failures may reflect infeasible foils or optimizer failure, as shown in Section~\ref{subsec:stress_test}.

Human studies are also needed, especially with participants outside MORL and RL, to test whether these explanations are understandable and useful. Prior RL-counterfactual work has evaluated whether non-expert users can identify flawed agents from counterfactual explanations, and whether counterfactual action-outcome visualizations improve users' understanding of agent preferences \cite{olson2021counterfactual,amitai2024coviz}. A similar study could test whether people can use command counterfactuals to distinguish well-trained agents from poorly trained agents in the same environment.

\section*{Acknowledgments}

This research has received funding from the project ALIGN4Energy (NWA.1389.20.251) of the research programme NWA ORC 2020 which is (partly) financed by the Dutch Research Council (NWO), and from the project PEER (grant agreement number 101120406) in the European Union’s Horizon Europe Research and Innovation Programme.

\bibliographystyle{named}
\bibliography{paper}

\appendix

\section{Appendix}
\label{app:appendix}

The Appendix is organized as follows:
\begin{itemize}
    \item Implementation Details
    \item \textsc{CF-ZOO} Hyperparameters
    \item Return-Only PCN Adjustment
    \item Landscape and Grid Construction
    \item Runtime Details
\end{itemize}

\subsection{Implementation Details}
\label{app:implementation_details}

\subsubsection{Scaled Command Coordinates}
\label{app:scaled_command_coordinates}

Because reward objectives can have different ranges and units, raw perturbation distances are not comparable. We therefore use scaled command coordinates. Let \(R_{\min}\) and \(R_{\max}\) be feasible command bounds and define
\begin{equation}
    D=\mathrm{diag}(\sigma),
    \qquad
    \sigma=R_{\max}-R_{\min}.
    \label{eq:appendix_scale_matrix}
\end{equation}
Invalid or near-zero ranges are replaced by \(1\).

Given a raw perturbation \(\delta\),
\begin{equation}
    z=D^{-1}\delta,
    \qquad
    \delta=Dz.
    \label{eq:appendix_scaled_coordinates}
\end{equation}
Equivalently,
\begin{equation}
    z_i
    =
    \frac{\delta_i}{R_{\max,i}-R_{\min,i}}.
\end{equation}
The scaled distance is
\begin{equation}
    \|z\|_2
    =
    \left\|D^{-1}\delta\right\|_2
    =
    \left(
    \sum_{i=1}^{d}
    \left(
    \frac{\delta_i}{R_{\max,i}-R_{\min,i}}
    \right)^2
    \right)^{1/2}.
    \label{eq:appendix_scaled_distance}
\end{equation}

Commands are projected to the feasible box
\begin{equation}
    \mathcal{C}
    =
    \{R:R_{\min}\leq R\leq R_{\max}\}.
    \label{eq:appendix_command_box}
\end{equation}
Thus, optimization evaluates
\begin{equation}
    R
    =
    \Pi_{\mathcal{C}}(R_t + Dz).
    \label{eq:appendix_projection}
\end{equation}

\subsubsection{Binary Refinement Along Successful Rays}
\label{app:binary_ray_refinement}

A successful endpoint may be farther than necessary, so we refine it by searching for the smallest successful step:
\begin{equation}
    \alpha_f(u)
    =
    \inf
    \left\{
    \alpha\in[0,\alpha_{\max}(u)]:
    M_f(R_t+\alpha D u)\geq\kappa
    \right\}.
    \label{eq:appendix_binary_ray_boundary}
\end{equation}

Binary search starts with
\[
    \alpha_{\mathrm{lo}}=0,
    \qquad
    \alpha_{\mathrm{hi}}=\alpha_{\max}(u).
\]
At each step,
\[
    \alpha_{\mathrm{mid}}
    =
    \frac{\alpha_{\mathrm{lo}}+\alpha_{\mathrm{hi}}}{2}
\]
is evaluated. If \(M_f(R_t+\alpha_{\mathrm{mid}}Du)\geq\kappa\), set \(\alpha_{\mathrm{hi}}=\alpha_{\mathrm{mid}}\); otherwise set \(\alpha_{\mathrm{lo}}=\alpha_{\mathrm{mid}}\).

We use \(16\) refinement steps. After \(K\) steps, the remaining interval is at most \(2^{-K}\) of the original interval. With \(K=16\),
\[
    2^{-16}\approx 1.5\times 10^{-5}
\]
of the original ray interval.

\subsubsection{ZOO-ADAM Coordinate Optimizer}
\label{app:zoo_adam_refinement}

This optimizer is a re-implementation of the ZOO-ADAM update rule from the ZOO attack \cite{chen2017zoo}.

The optimizer works in normalized coordinates:
\[
    z = D^{-1}\delta,
    \qquad
    \delta = Dz,
\]
where \(D=\mathrm{diag}(R_{\max}-R_{\min})\). Each policy query uses
\[
    R = \Pi_{\mathcal{C}}(R_t + Dz).
\]

At each iteration, coordinate \(i\in\{1,\ldots,d\}\) is sampled uniformly and evaluated at
\[
    z^{+}=z+h e_i,
    \qquad
    z^{-}=z-h e_i,
\]
where \(e_i\) is the \(i\)-th basis vector. In raw coordinates,
\[
    \delta^{+}=Dz^{+},
    \qquad
    \delta^{-}=Dz^{-}.
\]
The coordinate derivative is estimated by
\begin{equation}
    \widehat{\nabla_i \mathcal{L}}(z)
    =
    \frac{
        \mathcal{L}(z+h e_i)
        -
        \mathcal{L}(z-h e_i)
    }{2h}.
    \label{eq:appendix_zoo_finite_difference}
\end{equation}
We use \(h=10^{-3}\), requiring two policy queries.

The estimate updates coordinate-wise ADAM:
\begin{align}
    m_i^{(k)}
    &=
    \beta_1 m_i^{(k-1)}
    +
    (1-\beta_1)\widehat{\nabla_i \mathcal{L}}(z^{(k)}),
    \\
    v_i^{(k)}
    &=
    \beta_2 v_i^{(k-1)}
    +
    (1-\beta_2)
    \left(\widehat{\nabla_i \mathcal{L}}(z^{(k)})\right)^2.
\end{align}
We use
\[
    \beta_1=0.9,
    \qquad
    \beta_2=0.999,
    \qquad
    \epsilon=10^{-8}.
\]
Bias correction uses \(t_i\), the number of updates to coordinate \(i\):
\begin{equation}
    \hat{m}_i^{(k)}
    =
    \frac{m_i^{(k)}}{1-\beta_1^{t_i}},
    \qquad
    \hat{v}_i^{(k)}
    =
    \frac{v_i^{(k)}}{1-\beta_2^{t_i}}.
\end{equation}
The coordinate update is
\begin{equation}
    z_i^{(k+1)}
    =
    z_i^{(k)}
    -
    \eta
    \frac{
        \hat{m}_i^{(k)}
    }{
        \sqrt{\hat{v}_i^{(k)}}+\epsilon
    },
    \label{eq:appendix_zoo_adam_update}
\end{equation}
with \(\eta=0.01\). Other coordinates are unchanged.

After the update,
\[
    \delta^{(k+1)} = Dz^{(k+1)},
\]
and the PCN is queried at
\[
    R^{(k+1)}
    =
    \Pi_{\mathcal{C}}(R_t+\delta^{(k+1)}).
\]
The method returns the closest valid command found under
\[
    \left\|D^{-1}(R_{cf}-R_t)\right\|_2.
\]

\subsection{\textsc{CF-ZOO} Hyperparameters}
\label{app:ray_zoo_hyperparameters}

Table~\ref{tab:ray_zoo_hyperparameters} gives the \textsc{CF-ZOO} settings.

\begin{table}[ht]
\centering
\small
\setlength{\tabcolsep}{4pt}
\begin{tabular}{l r r}
\toprule
Parameter & White-box comparison & Stress test \\
\midrule
Random directions & \(21{,}000\) & \(9{,}001\) \\
Maximum queries & \(21{,}000\) & \(9{,}001\) \\
Ray binary-search steps & \(16\) & \(16\) \\
Margin \(\kappa\) & \(0.05\) & \(0.05\) \\
Penalty coefficient \(c\) & \(1.0\) & \(1.0\) \\
ZOO-ADAM learning rate & \(0.01\) & \(0.01\) \\
Finite-difference step \(h\) & \(10^{-3}\) & \(10^{-3}\) \\
Random seed & \(0\) & \(0\) \\
\bottomrule
\end{tabular}
\caption{\textsc{CF-ZOO} hyperparameters used in the reported experiments. The white-box comparison uses the same settings as the stress test except for the random-direction and query budgets.}
\label{tab:ray_zoo_hyperparameters}
\end{table}

\subsection{Return-Only PCN Adjustment}
\label{app:return_only_training_adjustment}

The original PCN formulation conditions the policy on both a desired return and a desired horizon. In our experiments, horizon conditioning sometimes introduced additional variation into the command-action mapping rather than consistently producing reliable temporal control. We therefore removed the horizon from the policy input, but retained its useful effect indirectly through trajectory filtering and relabelling.

First, when multiple trajectories achieved the same return vector, we preferred the shorter trajectory. This biases the replay buffer toward more efficient demonstrations for the same achieved outcome. Second, before relabelling trajectories, we removed reward-neutral suffixes. These are final parts of trajectories in which the agent no longer receives additional reward. Trimming them prevents unnecessary waiting or wandering behaviour from being treated as part of the intended command-conditioned behaviour. 

The return-only and return--horizon variants produced
comparable evaluation results and both learned useful multi-objective
behaviours. The return-only variant was therefore selected because it
preserved empirical performance while simplifying the command space.

\subsection{Landscape and Grid Construction}
\label{app:landscape_construction}

Each panel in Figure~\ref{fig:command_landscapes} shows how the policy's
decision changes when two components of the desired-return command are
varied. Everything else is kept fixed: the state \(s_t\), the foil action
\(a_f\), the trained policy, and all command components that are not shown
on the two axes.

Let \(p\) and \(q\) be the two displayed command dimensions. We select 121
equally spaced values for each dimension and evaluate every possible pair.
This produces
\[
121^2=14{,}641
\]
commands for each panel.

The policy is queried once for every command on this grid. We then calculate
the foil margin \(M_f(R)\). The colors in each panel show the value of
this margin, while the contour
\[
M_f(R)=\kappa
\]
marks the required validity threshold. Commands satisfying
\(M_f(R)\geq\kappa\) make the foil sufficiently preferred and are therefore
valid counterfactual commands.

Only two command dimensions can be displayed at once. Each panel is
therefore a two-dimensional slice through the full \(d\)-dimensional command
space. It shows what happens when the two displayed dimensions change while
all remaining dimensions stay fixed at their original values. It should not
be interpreted as a complete visualization of the full command space.

The feasibility audit in Table~\ref{tab:cfzoo_taxonomy} uses a different,
higher-dimensional grid. Instead of varying only two displayed dimensions,
it varies all \(d\) command dimensions. We use 200 equally spaced values for
each dimension and take their Cartesian product:
\[
\mathcal{G}
=
\mathcal{G}_1\times\cdots\times\mathcal{G}_d,
\qquad
|\mathcal{G}|=200^d.
\]
Thus, a two-dimensional command has \(200^2=40{,}000\) grid points, while a
three-dimensional command has \(200^3=8{,}000{,}000\) grid points.

A case is classified as grid-feasible if at least one evaluated command
satisfies
\[
\exists R\in\mathcal{G}
\quad\text{such that}\quad
M_f(R)\geq\kappa.
\]
This means that the grid search found at least one command that makes the
foil action valid.

\subsection{Runtime Details}
\label{app:runtime_details}

Table~\ref{tab:runtime_details} reports mean wall-clock time per
counterfactual method. \textsc{CF-ZOO} is slower because it adds directional
screening, refinement, and zeroth-order local search.

\begin{table}[H]
\centering
\small
\setlength{\tabcolsep}{3.2pt}
\begin{tabular*}{\columnwidth}{@{\extracolsep{\fill}}l l r@{}}
\toprule
Env. & Method & Time (s) \\
\midrule
Branch & \textsc{White-CW} & 9.44 \\
Branch & \textsc{CF-ZOO}  & 53.21 \\
B-Bottles & \textsc{White-CW} & 8.55 \\
B-Bottles & \textsc{CF-ZOO}  & 53.08 \\
Collect & \textsc{White-CW} & 7.98 \\
Collect & \textsc{CF-ZOO}  & 45.38 \\
DST & \textsc{White-CW} & 8.03 \\
DST & \textsc{CF-ZOO}  & 48.11 \\
Fruit & \textsc{White-CW} & 10.21 \\
Fruit & \textsc{CF-ZOO}  & 56.48 \\
Minecart & \textsc{White-CW} & 8.63 \\
Minecart & \textsc{CF-ZOO}  & 48.71 \\
R-Gather & \textsc{White-CW} & 11.37 \\
R-Gather & \textsc{CF-ZOO}  & 41.28 \\
R-Line & \textsc{White-CW} & 9.07 \\
R-Line & \textsc{CF-ZOO}  & 47.49 \\
\midrule
All & \textsc{White-CW} & 8.73 \\
All & \textsc{CF-ZOO}  & 48.78 \\
\bottomrule
\end{tabular*}
\caption{Mean runtime per counterfactual query. The aggregate row is weighted by the number of cases in each environment. Environment abbreviations follow the notation used in the main paper.}
\label{tab:runtime_details}
\end{table}

Table~\ref{tab:budget_runtime} reports mean and median wall-clock time per
query across budgets. Runtime grows steadily with the budget. We recommend
using 50k for the best compromise between results and time. 

Absolute
runtimes are hardware- and implementation-dependent and may be reduced
through faster hardware or further code optimization.

\begin{table}[H]
\centering
\footnotesize
\setlength{\tabcolsep}{1.8pt}
\begin{tabular*}{\columnwidth}{@{\extracolsep{\fill}}l r r@{}}
\toprule
Budget & Mean $\pm$ std (s) & Median (s) \\
\midrule
10k  &  19.1 $\pm$ 0.8 & 19.0 \\
20k  &  43.7 $\pm$ 1.8 & 43.5 \\
30k  &  60.5 $\pm$ 2.5 & 60.4 \\
50k  & 101.4 $\pm$ 2.6 & 102.1 \\
75k  & 108.4 $\pm$ 3.5 & 107.3 \\
100k & 176.0 $\pm$ 6.4 & 178.0 \\
\bottomrule
\end{tabular*}
\caption{Mean ($\pm$ standard deviation) and median wall-clock time per
counterfactual query across query budgets.}
\label{tab:budget_runtime}
\end{table}

\end{document}